\def\year{2021}\relax
\documentclass[letterpaper]{article} 
\usepackage{aaai21}  
\usepackage{times}  
\usepackage{helvet} 
\usepackage{courier}  
\usepackage[hyphens]{url}  
\usepackage{graphicx} 
\usepackage{natbib}  
\usepackage{caption} 
\usepackage{amssymb}
\usepackage{multirow}
\usepackage{siunitx}
\usepackage[switch]{lineno}
\usepackage{multirow}
\usepackage{varwidth}
\usepackage{diagbox}
\usepackage{amsmath}
\newcolumntype{P}[1]{>{\centering\arraybackslash}p{#1}}
\title{CARE: Commonsense-Aware Emotional Response Generation with Latent Concepts}
\author {
    Peixiang Zhong,\textsuperscript{\rm 1,2}
    Di Wang, \textsuperscript{\rm 2}
    Pengfei Li, \textsuperscript{\rm 3}
    Chen Zhang, \textsuperscript{\rm 4}
    Hao Wang, \textsuperscript{\rm 4}\thanks{Corresponding author}
    Chunyan Miao \textsuperscript{\rm 1,2}\footnotemark[1] \\
}
\affiliations {
    \textsuperscript{\rm 1} Alibaba-NTU Singapore Joint Research Institute, Nanyang Technological University (NTU), Singapore\\
    \textsuperscript{\rm 2} Joint NTU-UBC Research Centre of Excellence in Active Living for the Elderly, NTU, Singapore\\
    \textsuperscript{\rm 3} School of Electrical and Electronic Engineering, NTU, Singapore\\
    \textsuperscript{\rm 4} Alibaba Group, China \\
    \{peixiang001, wangdi, pli006, ascymiao\}@ntu.edu.sg, zhangchen010295@163.com, cashenry@126.com
    \\[3.0ex]
}
\begin{document}

\maketitle
\begin{abstract}
Rationality and emotion are two fundamental elements of humans. Endowing agents with rationality and emotion has been one of the major milestones in AI. However, in the field of conversational AI, most existing models only specialize in one aspect and neglect the other, which often leads to dull or unrelated responses. In this paper, we hypothesize that combining rationality and emotion into conversational agents can improve response quality. To test the hypothesis, we focus on one fundamental aspect of rationality, i.e., commonsense, and propose CARE, a novel model for commonsense-aware emotional response generation. Specifically, we first propose a framework to learn and construct commonsense-aware emotional latent concepts of the response given an input message and a desired emotion. We then propose three methods to collaboratively incorporate the latent concepts into response generation. Experimental results on two large-scale datasets support our hypothesis and show that our model can produce more accurate and commonsense-aware emotional responses and achieve better human ratings than state-of-the-art models that only specialize in one aspect.
\end{abstract}

\section{Introduction}
Rationality and emotion are two fundamental elements of humans and indispensable to our social interactions \cite{keltner1999social, colman2003cooperation}. Endowing agents with rationality and emotion has been one of the major milestones in AI. In recent studies of conversational AI, there is an emerging research trend in endowing conversational models with rationality \cite{zhou2018commonsense, zhang2020grounded} or emotion \cite{zhou2018emotional, song2019generating}. Rational conversational models can leverage commonsense knowledge to reason and have been shown to produce more appropriate and informative responses \cite{zhou2018commonsense}. Emotional conversational models can generate appropriate emotional responses, leading to improved user satisfaction \cite{prendinger2005empathic, zhou2018emotional} and long-term relationships with users \cite{zhou2018design}.

Rationality and emotion are not independent for humans \cite{de1990rationality}. In fact, emotions are often rational in social interactions \cite{pham2007emotion}. However, most existing conversational models only specialize in one aspect and neglect the other\footnote{One exception is XiaoIce \cite{zhou2018design}; however, it has no public API and only supports Mandarin.}, which often leads to dull or unrelated responses. For example, as shown in Table \ref{table: sample responses}, the state-of-the-art emotional conversational model (EmoDS) \cite{song2019generating} produces generic or unrelated emotional responses due to the lack of specific modelling of rationality. In addition, existing rational conversational models, e.g., the commonsense-aware conversational model (CCM) \cite{zhou2018commonsense}, are not able to generate emotional responses, rendering them difficult to build long-term relationships with users \cite{zhou2018design}. 
\begin{table}[!t]
\small
\centering
\begin{tabular}{p{1cm}|p{1cm}|p{5cm}}
 \hline
\multirow{2}{*}{\textbf{Model}}&\textbf{Message}& Why do you avoid starbucks?\\
\cline{2-3}
&\textbf{Gold}&I don't like their coffee.\\
 \hline
\multirow{3}{*}{EmoDS}&Neutral& I'm not a gamer so it's just that I'm a bit more tired of them.\\
\cline{2-3}
  &Joy&Yeah, that's pretty cool. I like that.\\
\cline{2-3}
  &Sadness&That sucks.\\
 \hline
\multirow{3}{1cm}{Ours (CARE)}&Neutral& It's just a general dislike of the \textbf{company}.\\
\cline{2-3}
  &Joy& \textbf{Tea} is my \textit{favorite}.\\
\cline{2-3}
  &Sadness&I'm not a big fan of \textbf{coffee}. It 's too \textit{bitter} for me. \\
\hline
\end{tabular}
\caption{Sample responses from EmoDS \cite{song2019generating} and our model. 
EmoDS generates generic or unrelated emotional responses. Our model extracts the message concept ``starbucks", and generates more commonsense-aware emotional responses by referring to our constructed relational (in bold) and emotional (in italic) latent concepts, e.g., company, coffee and bitter.}
\label{table: sample responses}
\end{table}

A recent work \cite{roller2020recipes} proposed to blend several human skills such as knowledge, personality, and empathy into a conversational agent and obtained the state-of-the-art performance in human evaluations. Their experimental analysis suggests that blending these skills is critical for achieving good human ratings. Motivated by the facts that 1) rationality and emotion are two fundamental qualities of humans and that 2) empirical performance improvement has been achieved via combining several human qualities \cite{roller2020recipes}, we hypothesize that combining rationality and emotion into conversational agents can improve response quality and their human ratings.

\begin{figure}[!t]
\centering
\includegraphics[width=0.95\linewidth]{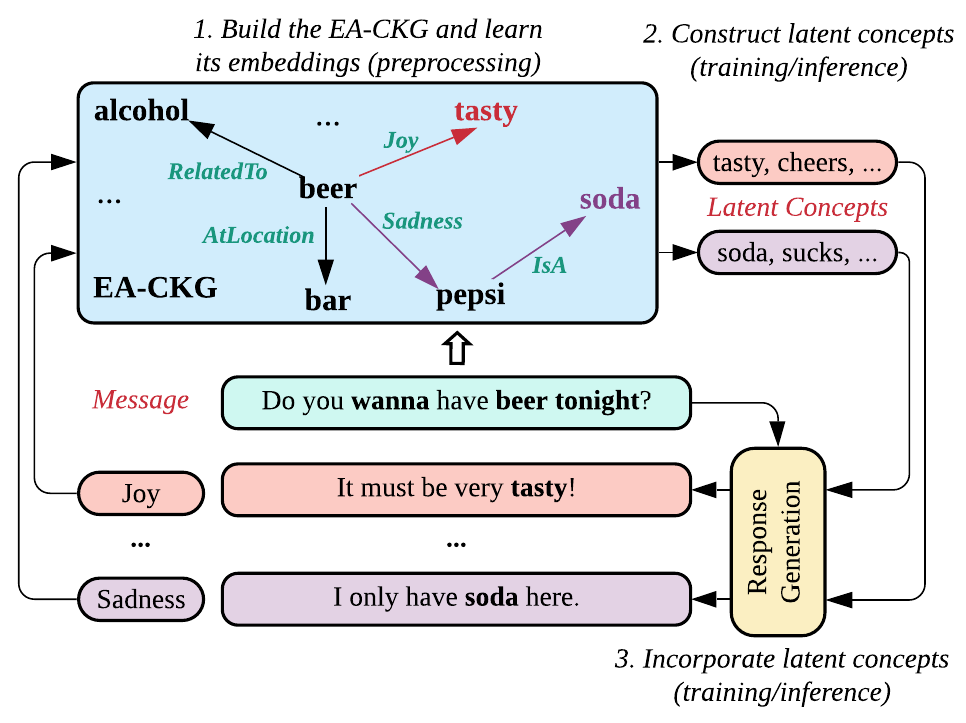}
\caption{Illustration of CARE. 
Given the message ``Do you wanna have beer tonight?" (``beer" is a message concept) and the learned EA-CKG embeddings, CARE first constructs latent concepts depending on the specified emotions of the response. For example, ``tasty" is constructed for ``joy" and ``soda" is constructed for ``sadness", because ``tasty" is linked to ``beer" via the ``joy" relation, and ``soda" is linked to ``beer" via a composite of ``sadness" and ``IsA" relations. Then CARE leverages the proposed three methods to incorporate the latent concepts, e.g., ``tasty", into response generation.}
\label{fig: illustration of CARE}
\end{figure}

In this paper, we narrow the scope of rationality and emotion to specific settings for easier implementation and evaluation. Specifically, we focus on one fundamental aspect of rationality, i.e., commonsense, and the discrete representation of emotion. 
Commonsense is an important foundation of rationality and the basis of rational human conversations \cite{ross1978rationality}. The discrete representation of emotion categorizes emotions into discrete basic emotions, e.g., joy, anger, etc., and is a well-established emotion theory in Psychology \cite{ekman1992argument}.
To test our hypothesis, we propose a novel model for \textbf{C}ommonsense-\textbf{A}ware \textbf{R}esponse generation with specified \textbf{E}motions (\textbf{CARE}) and assess its empirical performance.
Two major challenges to this task are 1) the lack of relevant datasets or resources that can provide such supervision and 2) how to generate appropriate commonsense-aware emotional words. We tackle the first challenge by building an emotion-aware commonsense knowledge graph (\textbf{EA-CKG}) to integrate commonsense and emotion knowledge. We tackle the second challenge by incorporating both relational and emotional latent concepts constructed from EA-CKG into response generation.
Specifically, we build EA-CKG by augmenting an external CKG with emotional triplets extracted from emotional conversations. We then construct latent concepts using learned EA-CKG embeddings, endowing the response with commonsense and emotion by reasoning over the EA-CKG. Finally, we propose three methods to sequentially and collaboratively incorporate the latent concepts during attention, optimization, and sampling. 
CARE is illustrated in Figure \ref{fig: illustration of CARE}. 

In summary, our contributions are as follows:
\begin{itemize}
    \item We identify the problem of lacking either rationality or emotion in existing conversational models, which often leads to dull or unrelated responses. We hypothesize that combining rationality and emotion into conversational agents can improve response quality.
    \item We focus on one fundamental aspect of rationality, i.e., commonsense, and propose CARE, the first commonsense-aware emotional response generation model, to address the aforementioned problem.
    \item We conduct extensive automatic and human evaluations and show that CARE can produce better commonsense-aware emotional responses than state-of-the-art models that only specialize in one aspect. The experimental results support our hypothesis.
\end{itemize}
\section{Related Work}
\noindent \textbf{Rational Response Generation}: Existing rational response generation models usually rely on knowledge bases, such as open-domain response generation \cite{han2015exploiting, young2018augmenting, ghazvininejad2018knowledge, liu2018knowledge, tuan2019dykgchat, moon2019opendialkg}, task-oriented response generation \cite{madotto2018mem2seq, wu2019globaltolocal} and question answering \cite{sun2018open, banerjee2019careful}. \citet{zhou2018commonsense} proposed CCM to incorporate commonsense knowledge by applying attention mechanisms on 1-hop knowledge triplets for open-domain response generation. \citet{zhang2020grounded} proposed ConceptFlow to extend CCM to multi-hop knowledge triplets. Different from CCM and ConceptFlow, our model is not restricted by the coverage of the CKG and can learn novel knowledge triplets for response generation.

\noindent \textbf{Emotional Response Generation}: 
Emotional conversational models \cite{hasegawa2013predicting, asghar2018affective, zhou2018mojitalk, zhong2019affect, rashkin2019towards, lin2019moel} are also emerging. \citet{zhou2018emotional} extended the Seq2Seq model by proposing an internal memory module to capture emotional state changes and an external memory module to generate emotional words. \citet{song2019generating} proposed an emotion classifier to guide the response generation. In contrast, our model generates emotional responses by leveraging emotional latent concepts constructed from KG embeddings. 

\noindent \textbf{Controlled Text Generation}: Recent controlled text generation methods are primarily based on generative adversarial networks (GAN) \cite{hu2017toward, li2019towards}, language models \cite{ghosh2017affect} and Seq2Seq models \cite{xing2017topic, xu2019neural}. \citet{keskar2019ctrl} trained a Transformer-based conditional language model on a large collection of corpora with control codes that govern style, content, and task-specific behavior. \citet{li2018syntactically} and \citet{peng2019topic} proposed topic-aware emotional response generation models. In contrast, we focus on commonsense, i.e., the semantic network of words, instead of topics, i.e., word clusters.

\section{Our CARE Model}
\label{sec: our approach}
In this section, we introduce the task definition and our CARE model, which includes a framework for constructing latent concepts and three methods to incorporate the latent concepts.

\subsection{Task Definition}
\label{sec: task definition}
We denote $\{X_i, Y_i, e_i\}, i=1, ..., N$, as a collection of \{\textit{message}, \textit{response}, \textit{emotion}\} tuples, where $e_i$ is chosen from a predefined set of emotions and denotes the emotion category of $Y_i$, and $N$ denotes the number of conversations in the training dataset. Our task can be formulated as follows: given a new message $X_{\text{new}}$ and an emotion category $e$, generate a natural and commonsense-aware response $Y_{\text{new}}$ that has emotion $e$. 

\subsection{Latent Concepts Construction Framework}
\label{sec: unified framework}
In this framework, we first build an emotion-aware commonsense knowledge graph (EA-CKG) and then construct latent concepts from EA-CKG.

\subsubsection{EA-CKG} 
We extract emotional triplets from emotional conversations and augment them into an external CKG to obtain EA-CKG. 
We use ConceptNet \cite{speer2017conceptnet} as our CKG\footnote{We remove non-English and rare concepts.}. Each triplet in ConceptNet follows the \{\textit{head}, \textit{relation}, \textit{tail}\} format, e.g., \{\textit{beer}, \textit{AtLocation}, \textit{bar}\}. 
Note that we use n-gram matching with ConceptNet to extract concepts from utterances, and ignore stopwords and n-grams that are formed entirely by stopwords. 
We define an emotional triplet as in the \{\textit{msg\_concept}, \textit{emotion}, \textit{res\_concept}\} format, representing an emotional link from a message concept to a response concept. For example, given a message ``I heard there is a bar nearby with nice beer.'' and its response ``I love tasty beer.'' with joy emotion, the triplet \{\textit{beer}, \textit{joy}, \textit{tasty}\} is a valid emotional triplet because there is a commonly expressed emotional link, i.e., \textit{joy}, from \textit{beer} in the message to \textit{tasty} in the response. 

We propose a two-step approach based on the pointwise mutual information (PMI) \cite{church1990word} to extract such emotional triplets from emotional conversations. PMI can measure the association between two words in a corpus. We extend the smoothed positive PMI, i.e.,  PPMI\textsubscript{$\alpha$} \cite{levy2015improving}, as follows:
\begin{equation}
\label{eqn: PPMI}
\text{PPMI}_{\alpha}(w_1, w_2)=\text{max} \left(\text{log} _{2} \frac{P(w_1, w_2)}{P_{\alpha}(w_1) P_{\alpha}(w_2)}, 0\right),
\end{equation}
where $(w_1,w_2)$ denotes the word pair, $P_{\alpha}(w) = \frac{\text{count}(w)^{\alpha}}{\sum_{x}{\text{count}(x)^{\alpha}}}$ denotes the smoothed probability of $w$, and $\alpha$ denotes a smoothing factor set to $0.75$ \cite{levy2015improving} to alleviate the bias towards rare words.
\begin{table}[!t]
\small
\centering
\begin{tabular}{cccc}
\hline
\textbf{CKG} & \textbf{\#entity} & \textbf{\#relation} & \textbf{\#triplet}\\
\hline
ConceptNet & 182K & 36 & 1.48M\\
EA-CKG (Reddit) & 182K & 42 & 1.58M\\
EA-CKG (Twitter) & 182K & 42 & 1.80M\\
\hline
\end{tabular}
\caption{EA-CKG statistics. Reddit and Twitter are two conversation datasets used in our experiments.}
\label{table: ea-ckg statistics}
\end{table}

In our two-step approach, we first construct a PPMI matrix between concepts in messages and in the corresponding responses to extract strongly associated concept pairs in conversations\footnote{We consider concept pairs whose frequency $\geq$ 5 and PPMI $\geq$ 1 as strongly associated pairs (CCP).}, denoted as conversational concept pairs (\textbf{CCP}). Note that in this case, $w_1$ refers to a message concept and $w_2$ refers to a response concept in Equation \ref{eqn: PPMI}. We then construct a second PPMI matrix between CCP and their expressed emotions and extract CCP that statistically express certain emotions more often than other emotions\footnote{We associate a CCP $\{w_1, w_2\}$ with emotion $e$ if PPMI($\{w_1, w_2\}, e$) $-$ $\max_{e_i \ne e}$PPMI($\{w_1, w_2\}, e_i$) $\geq$ 1.}. Note that in this case, $w_1$ refers to a CCP and $w_2$ refers to its expressed emotion in Equation \ref{eqn: PPMI}. We do not smooth $P(w_2)$. By using this two-step approach, we can effectively extract conversational triplets that are commonly expressed with certain emotions. The statistics of EA-CKG are presented in Table \ref{table: ea-ckg statistics}. Our approach shares similarities with commonsense knowledge base completion methods \cite{li2016commonsense, saito2018commonsense, bosselut2019comet}; however, they cannot be trivially adapted to extract emotional CCP.

\subsubsection{Latent Concepts Construction}
\label{sec: concept construction}
During training and inference, given a message $X_i$ and a desired emotion $e_i$, we construct the latent concepts of the response based on EA-CKG embeddings. Specifically, we first train a well-established knowledge embedding model, i.e., TransE \cite{bordes2013translating}\footnote{We adopt TransE because it achieves only marginally worse performance than RotatE \cite{sun2019rotate}, a state-of-the-art knowledge graph embedding model, for triplet classification on ConceptNet, but much faster in inference.}, on the entire EA-CKG to learn global concept and relation embeddings. The embeddings in TransE are learned such that the score -$||\mathbf{h}+\mathbf{r}-\mathbf{t}||_2$ for a correct triplet $(h, r, t)$ is much higher than a corrupted one, where $\mathbf{h,r,t}$ denote the TransE embeddings of $h, r, t$, respectively, and $||\mathbf{h}||_2=1$ and $||\mathbf{t}||_2=1$ \cite{bordes2013translating}. Hence, given a message concept $h$, a relation $r$ and a response concept $t$, we can estimate the relatedness between $h$ and $t$ via $r$ as follows:
\begin{equation}
\label{eqn: TransE score}
    \text{score}(h, r, t) = (\mathbf{h} + \mathbf{r})^{\top}\mathbf{t}.
\end{equation}
We then obtain the top $m$ related latent concepts of the response from EA-CKG, i.e., $\{t\}_1^m$, as follows:
\begin{equation}
\label{eqn: latent concept construction}
    \{t_i\}_{1}^{m} = \underset{t}{\mathrm{top}}(\text{score}(h, r, t)),
\end{equation}
where $h \in C_{X_i}$, $r \in R \cup \{e_i\}$, $C_{X_i}$ denotes all concepts in $X_i$, $R$ denotes all 36 relations in ConceptNet, and $t$ is searched over the concept vocabulary of EA-CKG. 
For messages without any concepts\footnote{Around 3\% messages do not have any concepts.}, we use a null message concept whose embedding is the average of all concept embeddings.

\subsubsection{Framework Analysis}
\label{sec: framework analysis}
Our framework constructs plausible relational ($r \in R$) and emotional ($r = e_i$) concepts for the response. By leveraging the EA-CKG embeddings, our framework inherits the ideas from knowledge base completion and has two major advantages over the graph search methods used in existing models \cite{zhou2018commonsense, zhang2020grounded} to find related concepts: 
1) our framework can find concepts that are both commonsense-aware and emotional due to the incorporation of emotional triplets in EA-CKG, e.g., \textit{tasty} is found given \textit{beer} and \textit{joy} whereas \textit{bland} is found given \textit{beer} and \textit{sadness}; and 2) our framework can not only traverse through the EA-CKG to find related concepts in a multi-hop neighborhood but also discover an arbitrary number of novel related concepts using Equation \ref{eqn: latent concept construction}, without being limited by the CKG coverage (see Result Analysis).

\subsection{Incorporating Latent Concepts}
\label{sec: incorporating latent concepts}
\begin{figure}[!t]
\centering
\includegraphics[width=0.85\linewidth]{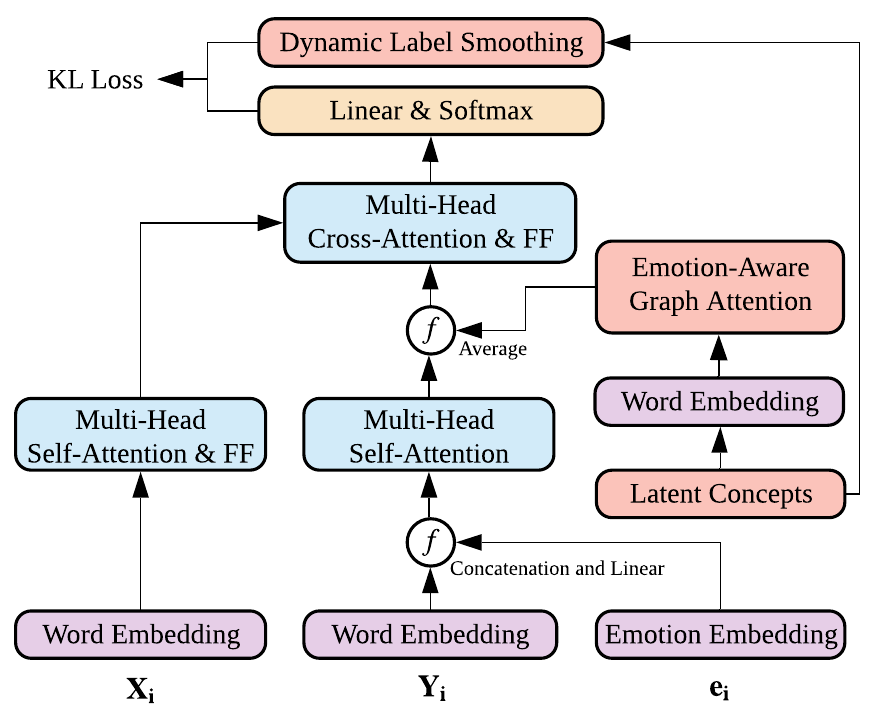}
\caption{Architecture of our Transformer-based conversational model. The positional encoding, residual connection, and layer normalization are omitted in the illustration for brevity.}
\label{fig: model}
\end{figure}
After obtaining the latent concepts, we propose three methods to collaboratively incorporate them into our Transformer-based conversational model \cite{vaswani2017attention}, as illustrated in Figure \ref{fig: model}. 
Note that similar to the idea of persona embedding \cite{li2016persona}, we additionally employ an emotion embedding layer in our decoder.

\subsubsection{Emotion-Aware Graph Attention}
We incorporate latent concepts into the decoder using an emotion-aware graph attention (\textbf{EAGA}) prior to the cross-attention layer, inspired by \cite{zhong2019knowledge}. We assume that important latent concepts are those related to the message concepts and have strong emotional intensity. The relatedness between concepts is obtained from Equation~\ref{eqn: TransE score}. The emotional intensity of a concept is computed based on an emotion lexicon NRC\_VAD \cite{mohammad2018obtaining} and an emotional intensity computation method \cite{zhong2019knowledge}.
We expand the size of NRC\_VAD from 20K to 34K using synonym expansion for better coverage\footnote{The expanded NRC\_VAD covers more than 97\% tokens in the datasets used in our experiments.}.

Formally, let $\{t_1, t_2, ..., t_m\}$ be the latent concepts of response $Y_i$ obtained from Equation \ref{eqn: latent concept construction}, $\{s_1, s_2, ..., s_m\}$ be their relatedness scores obtained from Equation~\ref{eqn: TransE score}, and $\{q_1, q_2, ..., q_m\}$ be their emotion intensities based on NRC\_VAD, we compute the latent concept embedding of $Y_i$, i.e., $\mathbf{C}_{Y_i}$, as follows:
\begin{equation}
\label{eqn: latent concept embedding}
    \mathbf{C}_{Y_i} = \sum_{i=1}^m{\beta_i \mathbf{t}_i},
\end{equation}
where $\mathbf{t}_i$ denotes the word embedding of $t_i$ and $\beta_i$ is computed as follows:
\begin{equation}
\label{eqn: beta}
    \beta_i = \lambda_i \frac{\exp(\delta_{1i} s_i)}{\sum_j{\exp(\delta_{1j} s_j)}} + (1-\lambda_i) \frac{\exp(\delta_{2i} q_i)}{\sum_j{\exp(\delta_{2j} q_j)}},
\end{equation}
where $\lambda_i$ denotes the trade-off coefficient between relatedness and emotional intensity, and $\delta_{1i}, \delta_{2i}$ denote softmax temperatures. Note that $\lambda_i$, $\delta_{1i}$ and $\delta_{2i}$ are concept-specific and can be fixed \textit{a prior} or learned during training. The obtained latent concept embedding $\mathbf{C}_{Y_i}$ is then averaged with the response representation prior to being fed to the cross-attention layer. Compared with the graph attention in CCM \cite{zhou2018commonsense}, EAGA measures concept relatedness using translation-based distance in TransE instead of MLP and additionally considers the emotion property of concepts.

\subsubsection{Dynamic Label Smoothing}
\label{sec: dynamic label smoothing}
Label smoothing is conventionally adopted in the Transformer \cite{vaswani2017attention} to improve translation quality. We propose a simple but effective dynamic label smoothing (\textbf{DLS}) method to explicitly enforce the supervision of latent concepts in producing concept-related responses, as well as to stabilize the learning process. Specifically, starting from the conventional label smoothing, we linearly increase the smoothing values for latent concepts with the training step and decrease the smoothing values for other words in the vocabulary. Note that the smoothing value of the target word remains unchanged. The maximum of the total smoothing value for latent concepts is a hyper-parameter to be tuned in experiments. 
We optimize model parameters to minimize the Kullback{-}Leibler (KL) loss \cite{kullback1951information}.
\subsubsection{Concept-Aware Top-$K$ Decoding}
\label{sec: concept-aware topk decoding}
During inference, we propose a concept-aware top-$K$ decoding (\textbf{CATD}) method to encourage the generation of words that are more related to the associated latent concepts. Formally, given the conventional top-$K$ unnormalized token probabilities $P(w_1), ..., P(w_k)$, our concept-aware token probability $P^{'}$ for $w_i, i=1, ..., k$, is computed as follows:
\begin{equation}
\label{eqn: CATD} 
    P^{'}(w_i) = P(w_i)*P_c^{\gamma}(w_i),
\end{equation}
where $\gamma$ denotes a trade-off hyper-parameter between fluency and relatedness, and $P_c(w_i)$ is computed as follows:
\begin{equation}
    P_c(w_i) = \frac{\exp(\mathbf{C}_Y^{\top} \mathbf{w}_i)}{\sum_{i=1}^{k} \exp(\mathbf{C}_Y^{\top}\mathbf{w}_i)},
\end{equation}
where $\mathbf{C}_Y$ denotes the latent concept embedding obtained from Equation~\ref{eqn: latent concept embedding} during inference. One merit of CATD is that it only reorders top-$K$ tokens by additionally considering their relatedness to latent concepts and thus does not introduce unlikely tokens into the sampling process. 

\section{Experimental Settings}
\label{sec: experimental settings}
In this section, we present the datasets, evaluation metrics, baselines, and model settings.
\subsection{Datasets}
\label{sec: datasets}
We conduct experiments on two large-scale datasets, namely Reddit and Twitter. The Reddit dataset is obtained from comments on the CasualConversation subreddit\footnote{https://www.reddit.com/r/CasualConversation/} discussing a variety of casual topics\footnote{https://files.pushshift.io/reddit/comments/}. The Twitter dataset is obtained from chats on twitter.com\footnote{https://github.com/Marsan-Ma/chat corpus/}. We truncate each sentence to a maximum of 30 tokens and use the most frequent 30K tokens as the vocabulary for each dataset.

\begin{table}[!t]
\small
\centering
\begin{tabular}{c|c|c|c}
\hline
 &  & \textbf{Reddit} & \textbf{Twitter}\\
\hline
\multirow{7}{*}{Training} & Neutral & 268K & 649K\\
& Joy & 232K & 308K\\
& Sadness & 236K & 302K\\
& Surprise & 551K & 543K\\
& Fear & 156K & 325K\\
& Anger & 132K & 373K\\
& Total & 1.58M & 2.50M\\
\hline
Validation & Total & 49K & 50K\\
\hline
Testing & Total & 49K & 50K\\
\hline
\end{tabular}
\caption{Dataset statistics.}
\label{table: datasets}
\end{table}

To obtain the ground-truth emotion label for each response, similar to \cite{zhou2018emotional, song2019generating}, we train an emotion classifier on emotional conversations. Specifically, we use the emotional tweets \cite{mohammad2012emotional, mohammad2018semeval} to train the classifier. We consider neutral and Ekman's six basic emotions \cite{ekman1992argument}: joy, sadness, surprise, fear, and anger, but exclude disgust due to its small amount of training samples in the emotional tweets. We propose an emotion classifier based on DeepMoji embeddings \cite{felbo2017using} followed by a linear layer and a softmax layer. Our classifier achieves an accuracy of 0.562 on a balanced test dataset, outperforming several competitive baselines such as BiLSTM (0.446), CNN (0.547), BERT \cite{devlin2019bert} (0.530) and XLNet \cite{yang2019xlnet} (0.522).
We then use the trained emotion classifier to annotate the responses in the datasets. The statistics of the annotated datasets are presented in Table \ref{table: datasets}.

\subsection{Evaluation Metrics}
\label{sec: evaluations}
We conduct both automatic and human evaluations. Automatic evaluation metrics include 1)~\textbf{Fluency}: perplexity (PPL), which measures the confidence of the generated responses; 2)~\textbf{Diversity}: distinct-1 (dist-1) and distinct-2 (dist-2) \cite{li2016diversity}, which measure the percentage of unique unigrams and bigrams in the generated responses, respectively; 3)~\textbf{Emotion Accuracy (EA)}: the emotion accuracy of the generated responses measured by our trained emotion classifier; and 4)~\textbf{Commonsense Awareness (CA)}: the average number of commonsense triplets in one pair of message and generated response, measured by ConceptNet. 

Following \cite{zhou2018emotional}, we conduct human evaluations to measure both \textbf{content quality} (rating scale in \{0, 1, 2\}) and \textbf{emotion quality} (rating scale in \{0, 1\}) of the generated responses. Content quality measures whether the response is natural and related to the message, as well as how commonsense-aware the response is. Emotion quality measures whether the response expresses the desired emotion appropriately and accurately. We randomly sample 200 test messages and emotions to generate 200 responses for each model. Each response is evaluated by three annotators. 

\subsection{Baselines}
\label{sec: baselines and model variants}
We compare CARE with the following baselines:

\noindent\textbf{Vanilla Models}: Seq2Seq \cite{vinyals2015neural} and Transformer \cite{vaswani2017attention}.

\noindent\textbf{Commonsense-Aware Models}: CCM \cite{zhou2018commonsense} and ConceptFlow \cite{zhang2020grounded}. ConceptFlow leverages multi-hop knowledge triplets and is a state-of-the-art model for commonsense-aware response generation.

\noindent\textbf{Emotional Models}: ECM \cite{zhou2018emotional} and EmoDS \cite{song2019generating}. EmoDS is a state-of-the-art model for emotional response generation.

\noindent\textbf{Pre-trained Model}: CTRL \cite{keskar2019ctrl}. CTRL is a large pre-trained conditional language model with 1.6 billion parameters trained on 140GB of text. We fine-tune CTRL on our training conversations such that it is able to produce emotional responses. CTRL has also been shown to contain commonsense knowledge \cite{petroni2019language, bosselut2019comet}.

\begin{table*}[!t]
\small
\centering
\begin{tabular}{c|ccccc|ccccc|cc}
\hline
 & \multicolumn{5}{c|}{\textbf{Reddit}} & \multicolumn{5}{c|}{\textbf{Twitter}} &\multirow{2}{*}{Size} & \multirow{2}{*}{IT}\\
\cline{0-10}
\textbf{Models} & PPL & Dist-1 & Dist-2 & EA & CA & PPL & Dist-1 & Dist-2 & EA & CA & &\\
\hline
Seq2Seq & \textbf{57.2} & 0.0035 & 0.0347 & - & 0.1349 & \textbf{79.7} & 0.0047 & 0.0522 & - & 0.1653 & 38M & \textbf{1.0x}\\
Transformer & 63.8 & 0.0032 & 0.0371 & - & 0.1224 & 90.1 & 0.0053 & 0.0563 & - & 0.1728 & \textbf{20M} & 1.5x\\
\hline
CCM & 62.3 & 0.0046 & 0.0469 & - & 0.1222 & 82.5 & 0.0060 & 0.0663 & - & 0.1835 & 74M & 5.9x\\
ConceptFlow & 60.1 & 0.0047 & 0.0458 & - & 0.1375 & 89.1 & 0.0051 & 0.0556 & - & 0.1893 & 33M & 21.8x\\
\hline
ECM & 65.6 & 0.0044 & \textbf{0.0506} & 0.5893 & 0.1105 & 91.3 & 0.0056 & 0.0630 & 0.5619 & 0.1650 & 40M & 2.0x\\
EmoDS & 76.6 & 0.0030 & 0.0455 & 0.6186 & 0.1107 & 113.5 & 0.0030 & 0.0450 & 0.5950 & 0.1599 & 46M & 1.5x\\
\hline
CTRL & - & \textbf{0.0068} & 0.0447 & 0.3425 & 0.1502 & - & \textbf{0.0108} & \textbf{0.0851} & 0.3995 & 0.1958 & 1.6B & 1876.7x\\
\hline
Ours (CARE) & 70.4 & 0.0049 & 0.0460 & \textbf{0.6840} & \textbf{0.1538} & 100.1 & 0.0064 & 0.0775 & \textbf{0.6693} & \textbf{0.2304} & 20M & 1.9x\\
\hline
\end{tabular}
\caption{Automatic evaluation results. Size denotes model size. IT denotes inference time relative to Seq2Seq.}
\label{table: automatic evaluation results}
\end{table*}

\begin{table*}[!t]
\small
\centering
\begin{tabular}{p{0.2cm}|p{1.8cm}|P{0.6cm}P{0.6cm}|P{0.6cm}P{0.6cm}|P{0.6cm}P{0.6cm}|P{0.6cm}P{0.6cm}|P{0.6cm}P{0.6cm}|P{0.6cm}P{0.6cm}|P{0.6cm}P{0.6cm}}
\hline
 & \multirow{2}{*}{\textbf{Models}} & \multicolumn{2}{c|}{Neutral} & \multicolumn{2}{c|}{Joy} & \multicolumn{2}{c|}{Sadness} & \multicolumn{2}{c|}{Surprise} & \multicolumn{2}{c|}{Fear} & \multicolumn{2}{c|}{Anger} & \multicolumn{2}{c}{Total}\\
\cline{3-16}
& & Cont & Emot& Cont & Emot& Cont & Emot& Cont & Emot& Cont & Emot& Cont & Emot& Cont & Emot\\
\hline
\multirow{4}{*}{\textbf{\rotatebox[origin=c]{90}{Reddit}}}& Seq2Seq & 0.62 & 0.34 & 0.79  & 0.32 & 0.69  & 0.15 & 0.78  & 0.35 & \textbf{0.72} & 0.19 & 0.74  & 0.08 & 0.73  & 0.24 \\
& ConceptFlow & 0.82 & 0.45 & 0.96 & 0.35 & 0.81 & 0.17 & 0.89 & 0.31 & 0.70 & 0.16 & 0.76 & 0.15 & 0.83 & 0.26\\
& EmoDS & 0.76 & 0.66 & 0.89 & 0.72 & 0.86 & \textbf{0.67} & 0.71 & 0.52 & 0.63 & 0.41 & 0.68 & 0.38 & 0.75 & 0.56\\
& CTRL & \textbf{0.92} & 0.50 & \textbf{1.08} & 0.63 & \textbf{1.03} & 0.42 & 0.79 & 0.34 & 0.66 & 0.24 & \textbf{0.93} & 0.38 & \textbf{0.90} & 0.42\\
& Ours (CARE) & 0.78 & \textbf{0.68} & 0.98 & \textbf{0.75} & 0.88 & 0.63 & \textbf{0.92} & \textbf{0.76} & 0.63 & \textbf{0.44} & 0.81 & \textbf{0.42} & 0.84 & \textbf{0.62}\\
\hline
\multirow{4}{*}{\textbf{\rotatebox[origin=c]{90}{Twitter}}}& Seq2Seq & 0.92 & 0.33 & 0.76 & 0.23 & 0.79 & 0.21 & 0.85 & 0.17 & 0.81 & 0.25 & 0.99 & 0.29 & 0.86 & 0.25 \\
& ConceptFlow & 0.97 & 0.42 & 0.91 & 0.28 & 0.98 & 0.22 & 1.03 & 0.19 & 0.87 & 0.21 & 0.85 & 0.26 & 0.93 & 0.27\\
& EmoDS & 0.82 & 0.46 & 0.78 & 0.48 & 0.91 & 0.56 & 0.93 & 0.63 & 0.79 & 0.65 & 0.84 & \textbf{0.65} & 0.84 & 0.57\\
& CTRL & \textbf{1.08} & 0.54 & \textbf{1.05} & \textbf{0.62} & \textbf{1.16} & 0.50 & \textbf{1.21} & 0.68 & 0.92 & 0.71 & \textbf{1.12} & 0.61 & \textbf{1.09} & 0.61\\
& Ours (CARE) & 0.87 & \textbf{0.57} & 0.83 & 0.58 & 1.13 & \textbf{0.62} & 1.15 & \textbf{0.71} & \textbf{0.93} & \textbf{0.74} & 0.94 & 0.63 & 0.96 & \textbf{0.64}\\
\hline
\end{tabular}
\caption{Human evaluation results. Cont and Emot denote content quality and emotion quality, respectively. The inter-annotator agreement, measured by Fleiss' Kappa \cite{fleiss1973equivalence}, are 0.441 and 0.626 for content and emotion on Reddit, respectively, and 0.479 and 0.673 for content and emotion on Twitter, respectively. Both datasets obtain ``moderate agreement" and ``substantial agreement" for content and emotion, respectively.}
\label{table: human evaluation results}
\end{table*}
\subsection{Model Settings}
\label{sec: model settings}
We use the same hyper-parameters for both datasets. 
Our TransE embeddings have a dimension of 100 and achieve an accuracy of 0.89 for triplet classification on EA-CKG.
Our Transformer model has 1 layer and 4 attention heads. We initialize the word embedding layer with pre-trained GloVe embeddings \cite{pennington2014glove} of size 300. The emotion embedding and feedforward layers have sizes of 50 and 512, respectively. We train our model using Adam \cite{kingma2014adam} with learning rate of 1, batch size of 64, and dropout of 0.1 for 80K steps, including 6K steps for warmup. We empirically construct 30 relational latent concepts and 10 emotional latent concepts for each response using Equation~\ref{eqn: latent concept construction}. We use label smoothing of 0.1, total smoothing value of 0.08 for latent concepts in DLS, and top-10 decoding with $\gamma=1$ in CATD. 

\section{Result Analysis}
\label{sec: result analysis}
In this section, we discuss our evaluation results, model analysis, case study, error analysis and limitation.

\subsection{Comparison with Baselines}
\label{sec: comparison with baselines}
We present the results of automatic evaluations in Table~\ref{table: automatic evaluation results}. Seq2Seq achieves the lowest perplexity while Transformer achieves slightly better diversity than Seq2Seq. Commonsense-aware models, i.e., CCM and ConceptFlow, obtain slightly better diversity and CA; however, they are unable to generate responses with specified emotions. Emotional models, i.e., ECM and EmoDS, achieve the highest EA among all baselines but the worst in perplexity and CA, suggesting that they only specialize in emotion and neglect commonsense. CTRL achieves the highest diversity among all models, partially due to its large vocabulary size of 250K. However, it obtains an inferior EA. Our model achieves better EA and CA than all baselines, including CTRL, which is also capable of producing commonsense-aware emotional responses. 

We present the results of human evaluations in Table \ref{table: human evaluation results}. The responses of non-emotional models are generated via top-10 decoding six times. ConceptFlow obtains similar emotion quality but noticeably better content quality than Seq2Seq due to its incorporation of multi-hop triplets. EmoDS achieves comparable content quality but much better emotion quality than Seq2Seq. CTRL obtains the best content quality among all models but only mediocre emotion quality, especially on Reddit. Our model performs best in emotion quality ($t$-test, $p<0.01$). In addition, our model achieves significantly better content quality than EmoDS ($t$-test, $p<0.01$), showing that our model can produce better commonsense-aware emotional responses than EmoDS. Finally, our model outperforms ConceptFlow, a competitive commonsense-aware model, in content quality, possibly because the graph search method in ConceptFlow heavily relies on the coverage of ConceptNet to extract knowledge triplets, but ConceptNet only has an average coverage of 27\% on Reddit and Twitter. In contrast, our model has less such restriction and can construct an arbitrary number of latent concepts given any input message.

We report model complexity in the rightmost columns of Table~\ref{table: automatic evaluation results}. Our model has comparable space and time complexity with vanilla baselines. In contrast, CTRL is around 80x larger and 1,000x slower than our model, rendering it intractable for real-time applications.
\begin{table*}[!t]
\small
\centering
\begin{tabular}{c|ccccc|ccccc}
\hline
 & \multicolumn{5}{c|}{\textbf{Reddit}} & \multicolumn{5}{c}{\textbf{Twitter}}\\
\hline
\textbf{Models} & PPL & Dist-1 & Dist-2 & EA & CA & PPL & Dist-1 & Dist-2 & EA & CA\\
\hline
Ours (CARE) & \textbf{70.4} & \textbf{0.0049} & 0.0460 & \textbf{0.6840} & 0.1538 & \textbf{100.1} & 0.0064 & 0.0775 & \textbf{0.6693} & 0.2304\\
\hline
-ET+EL & 72.2 & 0.0040 & 0.0428 & 0.6518 & 0.1332 & 100.8 & 0.0057 & 0.0669 & 0.6266 & 0.2077\\
-TransE& 72.8 & 0.0039 & 0.0430 & 0.6595 & 0.1261 & 101.8 & 0.0057 & 0.0660 & 0.6391 & 0.1960\\
-EAGA & 79.6 & 0.0045 & \textbf{0.0484} & 0.6258 & \textbf{0.1635} & 116.3 & \textbf{0.0080} & \textbf{0.1303} & 0.4775 & \textbf{0.3512}\\
-DLS & 72.6 & 0.0038 & 0.0441 & 0.6497 & 0.1277 & 100.7 & 0.0056 & 0.0682 & 0.6162 & 0.2050\\
-DLS + LS & 72.5 & 0.0040 & 0.0443 & 0.6421 & 0.1318 & 101.2 & 0.0055 & 0.0675 & 0.6194 & 0.2013\\
-CATD & \textbf{70.4} & 0.0036 & 0.0373 & 0.6094 & 0.1394 & \textbf{100.1} & 0.0059 & 0.0630 & 0.5848 & 0.1903\\
\hline
\end{tabular}
\caption{Ablation study. -ET+EL: replace the tails of the extracted emotional triplets (ET) by randomly sampled corresponding emotional words from an emotion lexicon (EL) \cite{mohammad13crowdsourcing}. -TransE: instead of using TransE, search neighbors with a growing neighborhood size (up to 3) on EA-CKG to find latent concepts. -EAGA: remove the emotion-aware graph attention. -DLS: remove the dynamic label smoothing. -DLS+LS: replace the dynamic label smoothing by conventional label smoothing (LS) of 0.1. -CATD: replace the concept-aware top-$K$ decoding by the conventional top-$K$ decoding.}
\label{table: ablation study}
\end{table*}
\begin{figure}[!t]
\centering
\includegraphics[width=\linewidth]{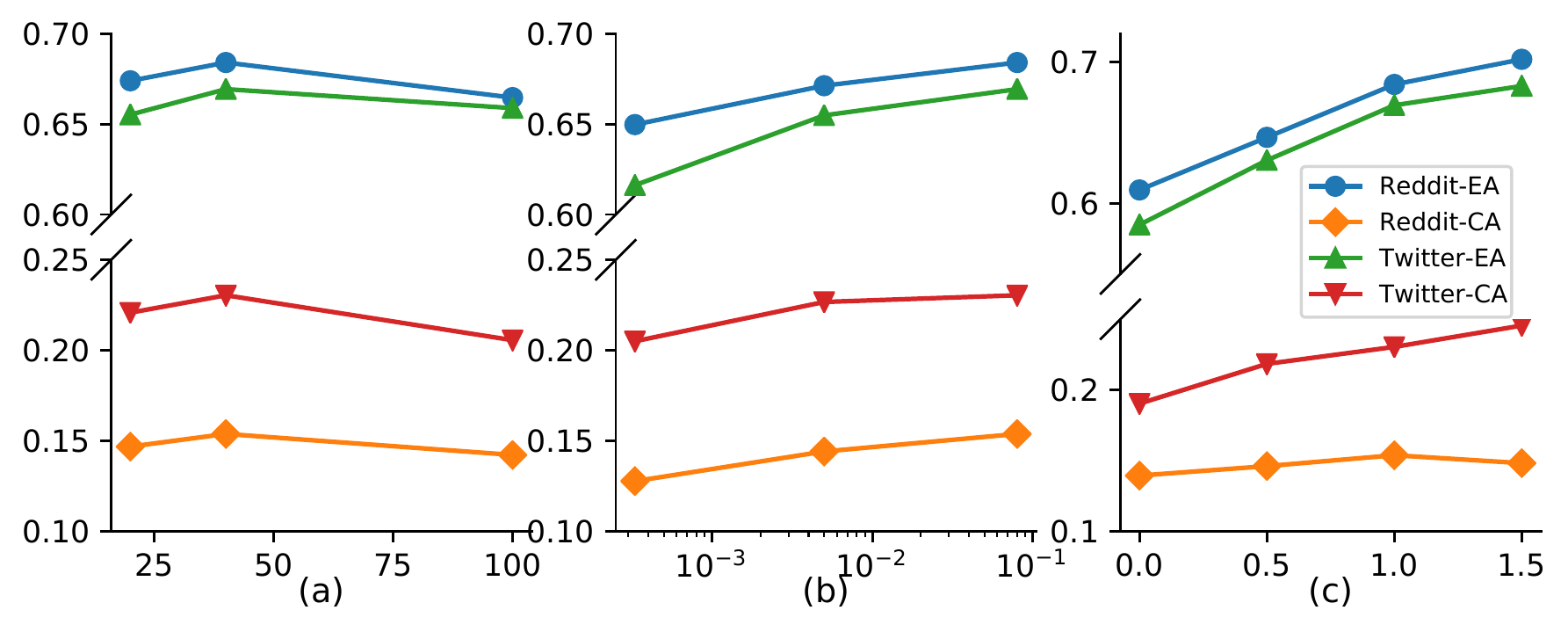}
\caption{Hyper-parameter analysis on EA and CA. (a)~Different number of latent concepts for each response (see $m$ in Equation \ref{eqn: latent concept construction}), where 1/4 latent concepts are emotional. (b)~Different total smoothing values for latent concepts in DLS. (c)~Different $\gamma$ (see Equation \ref{eqn: CATD}) in CATD.}
\label{fig: sensitivity analysis}
\end{figure}
\begin{table*}[!t]
\small
\centering
\begin{tabular}{c|c|P{7cm}|P{6cm}}
\hline
\multicolumn{2}{c|}{} & Why do you avoid starbucks? & I'd like to confess that I'm a cat person way more then a dog person.\\
\hline
\multicolumn{2}{c|}{Seq2Seq} & It's not bad. & I hope you get the chance.\\
\cline{1-4}
\multirow{6}{1cm}{Ours (CARE)} & Neutral &It's just a general dislike of the \textbf{company}. &I'm curious to hear the story behind this!\\
& Joy &\textbf{Tea} is my \textit{favorite}. &Yes! I \textit{love} their \textbf{eyes}, I \textit{love} them.\\
& Sadness &I'm not a big fan of \textbf{coffee}. It 's too \textit{bitter} for me. &I don't understand the difference though :(\\
& Surprise &I'm not sure. Maybe because of the \textbf{coffee shop} thing? &What kind of \textbf{cat} you like ?\\
& Fear &I'm not sure. It's just a \textit{weird} feeling. &That's the \textit{worst}!\\
& Anger &I hate the taste of \textbf{coffee}. It's just so \textit{bitter} and \textit{gross}. &That's a little \textit{annoying}!\\
\hline
\end{tabular}
\caption{Case studies. Words in bold and italic denote relational and emotional latent concepts, respectively.}
\label{table: case study}
\end{table*}
\subsection{Model Analysis}
\label{sec: model analysis}
We conduct ablation study, as shown in Table~\ref{table: ablation study}. Removing any component except EAGA from our model leads to much worse performance in both EA and CA. 
In particular, we observe that 1) our approach of constructing latent concepts performs better than alternatives (-ET+EL and -TransE); and 2) the removal of EAGA leads to significantly higher perplexity, diversity, and CA. 
The higher perplexity may be attributed to the additional supervisions of DLS on latent concepts, which are not explicitly incorporated into the model due to the lack of EAGA. 
The higher diversity and CA may be attributed to the untrained $\lambda$, $\delta_1$, and $\delta_2$ (see Equation \ref{eqn: beta}), which sometimes leads to ungrammatical but diverse latent concepts during decoding. 
Our observation validates the importance of EAGA in attending more related latent concepts.

We analyze the impact of model hyper-parameters on EA and CA, as shown in Figure~\ref{fig: sensitivity analysis}. Using $m=40$ latent concepts achieves the sweet spot for model complexity. Regarding DLS, increasing the total smoothing values for latent concepts in the $[0,0.08]$ range improves model performance. However, we do observe degraded fluency when using larger smoothing values, which is expected because the true learning signal is weakened. Increasing $\gamma$ in CATD consistently improves EA and CA for our model. However, models with larger $\gamma$, e.g., 1.5, sometimes produce unfluent long responses due to its overemphasizes on latent concepts.

\subsection{Case Study and Error Analysis}
\label{sec: case study}
We present two sample cases in Table~\ref{table: case study}. Given a message and desired emotions, our model produces commonsense-aware responses with the desired emotions, guided by both relational and emotional latent concepts.
For example, given ``starbucks" and anger, the relational latent concept ``coffee" and emotional latent concept ``gross" are constructed and incorporated into response generation. However, we do observe bad cases where the latent concepts overemphasize on emotional intensity, and the response becomes unnatural.
\subsection{Limitation}
\label{sec: limitation}
One major limitation of our work is the mediocre accuracy of our trained emotion classifier, which can be attributed to the unavailability of large-scale datasets for emotional conversations and sentences. Nevertheless, our proposed lightweight classifier obtains better performance than the best models reported in \cite{zhou2018emotional, song2019generating} and BERT. A potential solution to this limitation is to leverage few-shot learning on BERT-like models. 
\section{Conclusion}
\label{sec: conclusion}
We propose CARE as the first attempt to test the hypothesis that combing rationality (commonsense) and emotion into conversational agents can improve response quality and human ratings. 
Specifically, we build an EA-CKG and leverage its TransE embeddings to allow CARE to reason over the EA-CKG and construct both relational and emotional latent concepts. We further propose three methods to collaboratively incorporate the latent concepts into response generation.
Extensive ablation studies show that our methods of constructing and incorporating latent concepts outperform alternative methods. In addition, both automatic and human evaluations show that CARE can produce more accurate and commonsense-aware emotional responses than state-of-the-art commonsense-aware models and emotional models. 
Finally, our work provides empirical evidence for our hypothesis. In the future, we plan to extend our work to other aspects of rationality, e.g., logical reasoning.

\section*{Acknowledgments}
This research is supported, in part, by Alibaba Group through Alibaba Innovative Research (AIR) Program and Alibaba-NTU Singapore Joint Research Institute (JRI) (Alibaba-NTU-AIR2019B1), Nanyang Technological University, Singapore. This research is also supported, in part, by the National Research Foundation, Prime Minister's Office, Singapore under its AI Singapore Programme (AISG Award No: AISG-GC-2019-003) and under its NRF Investigatorship Programme (NRFI Award No. NRF-NRFI05-2019-0002). Any opinions, findings and conclusions or recommendations expressed in this material are those of the authors and do not reflect the views of National Research Foundation, Singapore. This research is also supported, in part, by the Singapore Ministry of Health under its National Innovation Challenge on Active and Confident Ageing (NIC Project No. MOH/NIC/COG04/2017 and MOH/NIC/HAIG03/2017).
\bibliography{aaai21}
\end{document}